\documentclass[10pt]{article} 
\usepackage[preprint]{rlc}

\usepackage{amssymb}            
\usepackage{mathtools}          
\usepackage{mathrsfs}           
\usepackage{graphicx}           
\usepackage{subcaption}         
\usepackage[space]{grffile}     
\usepackage{url}                
\usepackage{algorithm}
\usepackage{algpseudocode}
\usepackage{xcolor}

\DeclareMathOperator{\E}{\mathbb{E}}

\newcommand\nnfootnote[1]{%
  \begin{NoHyper}
  \renewcommand\thefootnote{}\footnote{#1}%
  \addtocounter{footnote}{-1}%
  \end{NoHyper}
}

\title{Robust Model-Based Reinforcement Learning with an Adversarial Auxiliary Model}

\author{Siemen Herremans  \\
    siemen.herremans@uantwerpen.be \\
    IDLab, University of Antwerp - imec \\
    \And
    Ali Anwar \\
    ali.anwar@uantwerpen.be\\
    IDLab, University of Antwerp - imec \\
    \And
    Siegfried Mercelis \\
    siegfried.mercelis@uantwerpen.be\\
    IDLab, University of Antwerp - imec}

\graphicspath{{Images/}}

\begin{document}

\maketitle

\nnfootnote{RL Safety Workshop (RLSW) @ RLC 2024}

\begin{abstract} 

Reinforcement learning has demonstrated impressive performance in various challenging problems such as robotics, board games, and classical arcade games. However, its real-world applications can be hindered by the absence of robustness and safety in the learned policies. More specifically, an RL agent that trains in a certain Markov decision process (MDP) often struggles to perform well in nearly identical MDPs. To address this issue, we employ the framework of Robust MDPs (RMDPs) in a model-based setting and introduce a novel learned transition model. Our method specifically incorporates an auxiliary pessimistic model, updated adversarially, to estimate the worst-case MDP within a Kullback-Leibler uncertainty set. In comparison to several existing works, our work does not impose any additional conditions on the training environment, such as the need for a parametric simulator. To test the effectiveness of the proposed pessimistic model in enhancing policy robustness, we integrate it into a practical RL algorithm, called Robust Model-Based Policy Optimization (RMBPO). Our experimental results indicate a notable improvement in policy robustness on high-dimensional MuJoCo control tasks, with the auxiliary model enhancing the performance of the learned policy in distorted MDPs. We further explore the learned deviation between the proposed auxiliary world model and the nominal model, to examine how pessimism is achieved. By learning a pessimistic world model and demonstrating its role in improving policy robustness, our research contributes towards making (model-based) RL more robust. 

\end{abstract}

\section{Introduction}
\label{sec:introduction}

Reinforcement learning (RL) has been shown to perform well in many environments. However, the performance of a trained RL agent can rapidly decrease when the agent is evaluated in a slightly altered environment \citep{christiano2016transfer, rusu2017sim}. This is one of the issues that has limited the adoption of RL in real-world scenarios, more specifically due to the simulation-to-reality (sim2real) gap and inherent variability in real control systems. Therefore, there is a need for policies that are robust enough to perform well in environments that slightly differ from the training environment. Due to this necessity, various approaches tackle the sim2real issue, often using different problem formulations \citep{zhao2020sim}. Some of these approaches include domain randomization or transfer learning. In our work, the goal is to maximize the worst-case performance of the RL agent, commonly formalized as a \emph{Robust Markov decision processes} (RMDP). This formalism defines an \emph{uncertainty set} of multiple MDPs, where the agent is oblivious to which MDP it is acting in. The objective in an RMDP then becomes to maximize the return in the worst (lowest cumulative reward) MDP of the uncertainty set. In previous research, methods that work within the RMDP formals have demonstrated enhanced robustness against perturbations between the train and test environment \citep{wang2024bring, pinto2017robust}. However, these works often impose extra requirements on the training environment, such as the ability to re-sample a transition multiple times or to have access to a parametric environment during training. This paper follows the RMDP formulation and proposes a novel algorithm that improves the robustness of a learned policy, without placing any additional requirements on the training environment. Inspired by the ideas of \cite{rigter2022rambo} and \cite{pinto2017robust}, our approach introduces an auxiliary model that acts as an adversary to minimize the cumulative reward under the current policy. This auxiliary model's objective then acts in a two-player Markov game with the policy optimization objective. By sequentially optimizing these two competing objectives, our algorithm can optimize towards a more robust policy. Our \textbf{main contributions} are firstly \emph{(i)}, proposing a novel robust model-based RL algorithm to improve robustness in an online setting. This is achieved by adding an auxiliary model to MBPO which learns a pessimistic world model via adversarial updates. Secondly \emph{(ii)}, we evaluate the empirical performance of our algorithm on high-dimensional Gym MuJoCo control benchmarks \footnote{Evaluation code and weights available at \url{https://github.com/rmbpo-eval/rmbpo-eval}}. Thirdly \emph{(iii)}, we analyze how the predictions of the learned robust model differ from the nominal model. The remainder of this work will first describe the relevant background to our approach. Then, the methodology is described in detail. Subsequently, the results demonstrate the improvement in robustness that our method provides to MBPO \citep{janner2019trust} in multiple MuJoCo \citep{todorov2012mujoco} control environments. Finally, we draw conclusions and outline future research directions.

\section{Background}
\label{sec:preliminaries}

In this section, we first introduce model-based RL (MBRL) within the broader context of Markov decision processes (MDP). Secondly, RMDPs are described and an adversarial framework to tackle them is highlighted. Finally, the Kullback-Leibler (KL) uncertainty set is defined. 

\subsection{Model-Based Reinforcement Learning}
\label{sec:mbrl}

Model-Based Reinforcement Learning (MBRL) \citep{moerland2023model} operates within the framework of a Markov decision process (MDP), defined by the tuple $(\mathcal{S}, \mathcal{A}, P, r, \gamma, \rho_0)$, where $\mathcal{S}$ and $\mathcal{A}$ denote the state and action spaces, $P(s'|s, a)$ is the distribution that defines the probability of ending up in next state $s'$ when taking action $a$ in state $s$. Next, $r(s, a, s')$ defines the reward function, $\gamma$ is the discount factor, and $\rho_0(s)$ is the initial state distribution. The objective in RL is to identify an optimal policy $\pi^*$ that maximizes the expected sum of discounted rewards:

\begin{equation}
    \pi^* = \arg\max_\pi \mathbb{E}_{\pi, P, \rho_0} \left[ \sum_{t=0}^H \gamma^t r(s_t, a_t, s_{t+1}) \mid  s_0 \sim \rho_0 \right]
\end{equation}

In addition, we denote the state visitation distribution of the MDP as $d^\pi_P$, which defines the likelihood of being in a state certain state when following policy $\pi$. In MBRL, the agent learns a model of the environment's dynamics, represented by $p_\theta(s'|s, a)$, from the data collected through its interactions with the MDP. This model is then used to simulate future states and rewards, reducing the number of interactions with the real environment. The expected reward function, $r(s, a)$, is also learned from data. In most MBRL algorithms, the agent's policy is updated based on both real experiences and simulated experiences from the learned model, balancing between exploration for model learning and exploitation of the learned model for policy improvement. For notational simplicity, we will use $s$, $a$ and $s'$ to denote $s_t$, $a_t$, $s_{t+1}$ respectively, when it is clear from context.

\subsection{Robust Markov Decision Processes}
\label{sec:rmdp}

In a traditional MDP, the agent optimizes its policy in a static transition model $P$. However, in some real-world problems, the transition model can change over time. Hence, we can define a Robust MDP \citep{wiesemann2013robust} where the agent acts in an unknown MDP $P \in \mathcal{P}$ that is a sample from an uncertainty set $\mathcal{P}$. The robust objective $J_{\mathcal{P}, \pi}$ can now be defined to maximize an objective function in the worst-case MDP of a given uncertainty set. This objective is formally stated in Eq.~\ref{eq:objective}.

\begin{equation}    
    J_{\mathcal{P}, \pi} = \max_{\pi \in \Pi} \min_{P \in \mathcal{P}} \E_{P, \pi} \left[ \sum_{t=0}^H \gamma^t r(s_t, a_t, s_{t+1}) \mid  s_0 \sim \rho_0 \right]
    \label{eq:objective}
\end{equation}

The optimal policy ($\pi^*_{\mathcal{P}}$) now becomes the policy that maximizes $J_{\mathcal{P}, \pi}$ (over the set of achievable policies $\Pi)$, this is called the outer-loop problem. Additionally, the algorithm is dependent on knowing the worst-case MDP at every time step, we call this the inner-loop problem. For a small uncertainty set, the inner-loop problem can be solved by just evaluating a certain transition in each MDP $P \in \mathcal{P}$. However, when the uncertainty set becomes very large or continuous, the inner-loop problem can be challenging. We will follow related works by considering this combined optimization objective as a two-player zero-sum Markov game \citep{rigter2022rambo, pinto2017robust}. In this game, one player optimizes the policy, to maximize the return, whilst the other player tries to find $P^* \in \mathcal{P}$, which minimizes the return. Both these players are updated in an alternating manner.

\subsection{KL Uncertainty set}
\label{sec:kl_uncertainty}

Since the "true" uncertainty set is often not known or ill-defined, a common choice is the Kullback-Leibler (KL) uncertainty set, denoted as $\mathcal{P}_{KL}$. The KL uncertainty set is defined as:

\begin{equation}
    \mathcal{P}_{KL} = \left\{ P \in \mathcal{P}_{feasible} \; | \; D_{KL}(P || \bar{P}) \leq \epsilon \right\},
\end{equation}

where $\bar{P}$ is the nominal kernel, i.e. the environment with which the agent interacts during training. $\mathcal{P}_{feasible}$ denotes the set containing all MDPs under consideration, in the case of a parametric model, $\mathcal{P}_{feasible}$ contains every MDP that can be represented by that model. $D_{KL}(P || \bar{P})$ is the KL divergence between the model $P$ and the nominal model $\bar{P}$, and $\epsilon$ is a predefined threshold. In this definition, the KL uncertainty set $\mathcal{P}_{KL}$ consists of all models that are within a KL divergence of $\epsilon$ from the nominal model $\bar{P}$. This set is the uncertainty set that will be approximated in our work.

\section{Auxiliary Adversarial Model}
\label{sec:aux_model}

The goal of this section is to tackle the inner-loop problem of the robust objective, as defined by the minimization problem in Eq.~\ref{eq:objective}, i.e. approximating the worst-case MDP, denoted as $P^* \in \mathcal{P}$, where we choose $\mathcal{P}$ to be the KL uncertainty set centered around the nominal model $\bar{P}$. This choice of uncertainty set follows a common choice in literature \citep{wang2024bring, hu2013kullback}. To describe our methodology, this section first (Section \ref{sec:aux_model}) introduces the auxiliary adversarial model as an addition to traditional world model learning (e.g. via maximum likelihood estimation \citep{janner2019trust}). The auxiliary model has a well-defined KL divergence with the approximated nominal model. Secondly (Section \ref{sec:training_aux_model}), we introduce the loss function to train the auxiliary model to maintain a low KL divergence with the nominal transition model, whilst also learning to be pessimistic (i.e., minimizing the return of the transition). Finally, we propose \textbf{Robust MBPO (RMBPO)}, an algorithm that incorporates the auxiliary model to improve the robustness of the learned policy.

\subsection{Auxiliary Model} \label{sec:aux_model}

Since we work within the context of Model-Based RL, we have direct access to a parameterized approximation, $p_{\theta}(s', r | s, a)$, of the nominal transition model $\bar{P}(.)$. However, this does not directly provide us with a method to approximate $D_{KL}(p_\theta || \bar{P})$, since we do not have access to the simulated transition probabilities, $\bar{P}(s', r | s, a)$, needed to construct the KL uncertainty set. Hence, we propose to not directly try to approximate the pessimistic transition model, thus leaving $p_{\theta}$ untouched. As an alternative, we propose an auxiliary parameterized model, $g_\psi$, which takes as input the outputs of the learned transition model $p_\theta$, in addition to $s$ and $a$. Next states and rewards can now be sampled according to Eq.~\ref{eq:auxiliary}.

\begin{equation}
    s', r \sim g_\psi(\cdot \, | \, s, a, p_{\theta}(s', r|s, a))
    \label{eq:auxiliary}
\end{equation}

Since both $p_\theta$ and $g_\psi$ define probability distributions, it is possible to compute $D_{KL}(g_\psi || p_\theta)$, which we will consider as an approximation for $D_{KL}(g_\psi || \bar{P})$. In our work, both $p_{\theta}$ and $g_\psi$ define the mean and covariance matrix of a diagonal multivariate Gaussian distribution, so the KL divergence can be computed closed-form. In practice, we provide the predicted mean $\mu_\theta$ and covariance matrix $\Sigma_\theta$ as inputs to the auxiliary model $g_\psi$, since a Gaussian is fully defined by these two components. Strictly speaking, the addition of $p_\theta$ as an input to the auxiliary model is not necessary, however, this greatly eases the optimization of $g_\psi$, which will be explained in Section \ref{sec:training_aux_model}.

\subsection{Training the Auxiliary Model} \label{sec:training_aux_model}

The goal of the auxiliary model is to minimize the value under the current policy while remaining within the desired uncertainty set $\mathcal{P}_{KL}$. To incentivize both these objectives, the auxiliary model is trained by minimizing a sum of two loss functions, displayed in Eq.~\ref{eqn:aux_loss}. The first term is the Kullback-Leibler (KL) divergence between the normal model $p_\theta$ and the auxiliary model $g_\psi$, which ensures that the auxiliary model does not deviate too far from the (approximated) dynamics of the nominal environment. Therefore, this part of the loss function incentivizes the model to remain within the KL uncertainty set $\mathcal{P}_{KL}$. The second term of the loss function minimizes the return of the current state transition, as proposed by \cite{rigter2022rambo}. By lowering the (log) likelihood of transitions with a high return and heightening the likelihood of transitions with a low return, the auxiliary model will become a pessimistic approximation of $p_\theta$ and $\bar{P}$.

\begin{equation}
J_g(\psi) = \mathbb{E}_{(s', r) \sim g_\psi, s \sim d^\pi_{\psi, \theta}, a \sim \pi} \left[ D_{KL}(g_{\psi} || p_\theta) + \eta \cdot \ln(g_{\psi}(s', r| s,a,p_\theta(\cdot | \cdot)) \cdot \textbf{sg}(r + \gamma V^{\theta, \phi}_{\psi}(s')) \right]
\label{eqn:aux_loss}
\end{equation}

In Eq.~\ref{eqn:aux_loss}, \(\textbf{sg()}\) defines the \emph{stop\_grad} operator, no gradient is computed for the value, only for \(g_\psi\). $V(.)$ denotes the (approximate) value function, used to estimate the expected return after the transition. The hyperparameter $\eta$ controls the influence of the value function: for a small $\eta$, $p_\theta \approx g_\psi$ and therefore $\mathcal{P}_{KL}$ is small, for larger values of $\eta$, $g_\psi$ will grow more pessimistic and therefore $\mathcal{P}_{KL}$ can be large. The values of $\eta$ that were used in this work can be found in Appendix \ref{appendix:hyperparams}.

\subsection{Proposed Algorithm}

To improve policy robustness, we combine the auxiliary model with MBPO \citep{janner2019trust} to become RMBPO. MBPO approximates the training environment by maximizing the likelihood of experienced transitions under its learned model $p_\theta$. This model is a neural network that predicts a mean and covariance matrix over the next states and rewards, conditioned on the current state and action. On-policy rollouts are then performed on the learned model. Finally, the unrolled data is used to update a policy via Soft-Actor Critic (SAC) \citep{haarnoja2018soft}. We modify MBPO by training an auxiliary model in addition to the existing model, via Eq.~\ref{eqn:aux_loss}. Since these two models are trained separately, the auxiliary model learning does not hinder the accuracy or precision of $p_\theta$. During the model unroll, we pass the current state through the learned model $p_\theta$, after which we use the output of that model ($\mu_\theta, \Sigma_\theta)$ as input to the auxiliary model. The auxiliary model will then predict a modified ($\mu_\psi, \Sigma_\psi)$ as an approximation to the worst-case transition model in $\mathcal{P}_{KL}$. Relating to Section \ref{sec:rmdp}, the auxiliary model tries to solve the inner-loop problem, while SAC tries to maximize the outer-loop problem. These two components act as two players in a zero-sum Markov game. This algorithm is fully described in Algorithm \ref{alg:ram}, where our additions are highlighted in blue. Following other works \citep{wang2024bring, zhou2024natural}, we add a small amount of action noise to the environment, otherwise, the uncertainty set would not be well-defined. More details on the action noise are provided in Appendix \ref{appendix:details}.

\begin{algorithm}
\caption{RMBPO \textcolor{blue}{(Additions in blue)}}\label{alg:ram}
\begin{algorithmic}[1]
\State{Initialize policy $\pi_{\phi}$, predictive model $p_\theta$ , \textcolor{blue}{auxiliary model $g_\psi$}, \\ environment dataset $\mathcal{D}_{env}$, model dataset $\mathcal{D}_{model}$}

\For{N epochs}

\While {improving}
\State{Update model parameters $\theta$ on environment data: $\theta \leftarrow \theta - \lambda_p \hat{\nabla}_\theta J_p(\theta, \mathcal{D}_{env})$}
\EndWhile
\textcolor{blue}{
\While {improving}
\State{Update model parameters $\psi$ according to Eq.~\ref{eqn:aux_loss}: $\psi \leftarrow \psi - \lambda_a \hat{\nabla}_\psi J_g(\psi, \mathcal{D}_{env}, p_\theta, \pi_\phi$)}
\EndWhile
}
\For{E steps}
\State{Take action in environment according to $\pi_{\phi}$; add to $\mathcal{D}_{env}$}
\For{M model rollouts}
\State{Sample $s_t$ uniformly from $\mathcal{D}_{env}$}
\State{On-policy \textcolor{blue}{rollout according to Eq.~\ref{eq:auxiliary}} starting from $s_t$ using policy $\pi_{\phi}$; add to $\mathcal{D}_{model}$}
\EndFor
\State{Perform (soft) actor-critic updates on $\phi$ using samples from ${D}_{model}$.}
\EndFor
\EndFor
\end{algorithmic}
\end{algorithm}

\section{Results}
\label{sec:results}

The following section aims to answer two main research questions: \emph{"Does the auxiliary model learn pessimistic state transitions?"} and \emph{"Can the auxiliary model make a learned policy more robust?"}. The first question is investigated in Section \ref{subsec:model_results}, where we perform a case study on the Hopper-v3 environment to examine which changes are made by the auxiliary model. The second question is investigated in Section \ref{subsec:main_results}, where we plot the performance of RMBPO compared to MBPO under distorted evaluation environments.

\subsection{Main Results}
\label{subsec:main_results}

\begin{figure}
    \centering
    \begin{subfigure}{0.49\textwidth}
        \includegraphics[width=\textwidth]{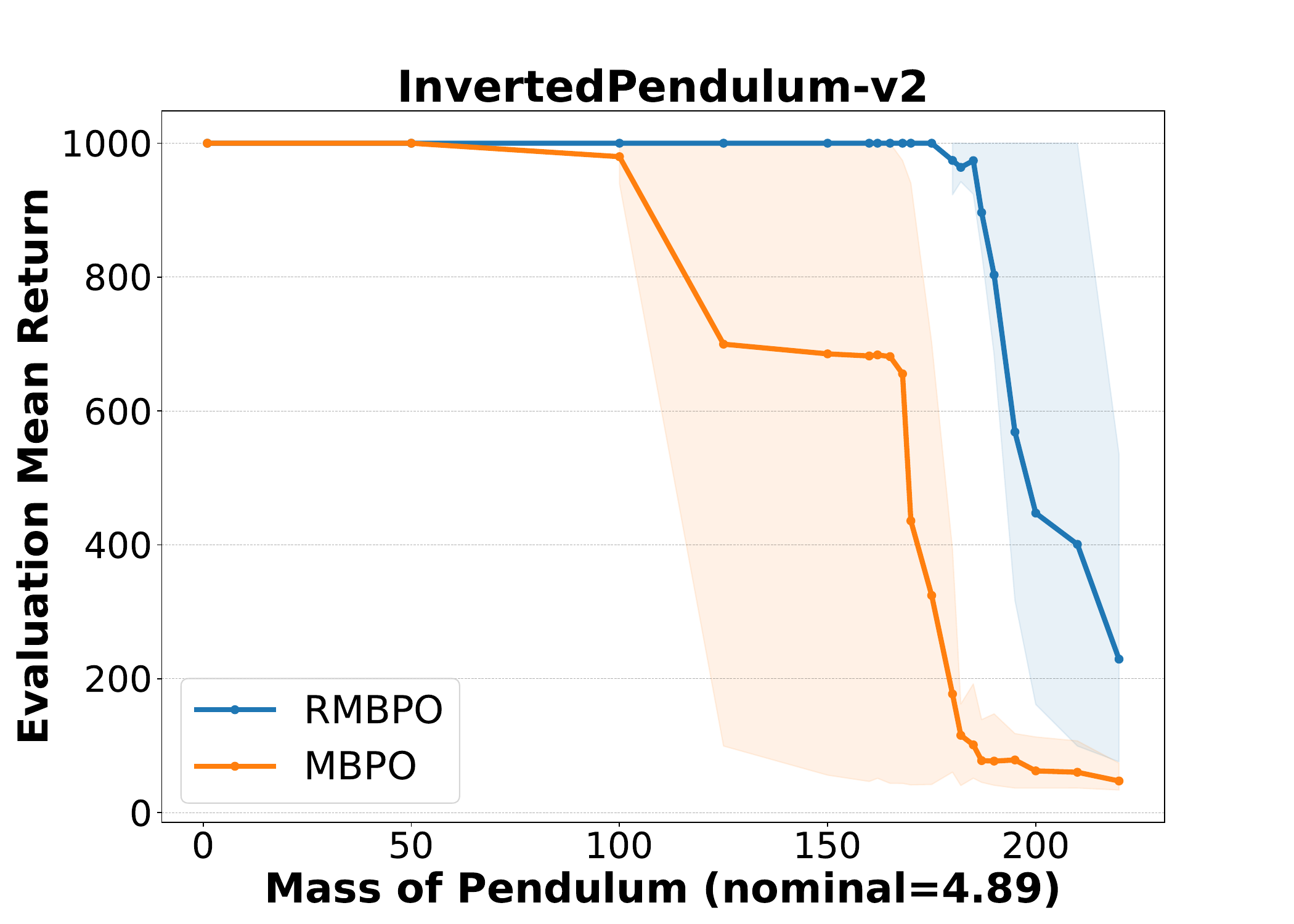}
        \caption{}
        \label{fig:pendulum_robustness1}
    \end{subfigure}
    \hfill
    \begin{subfigure}{0.49\textwidth}
        \includegraphics[width=\textwidth]{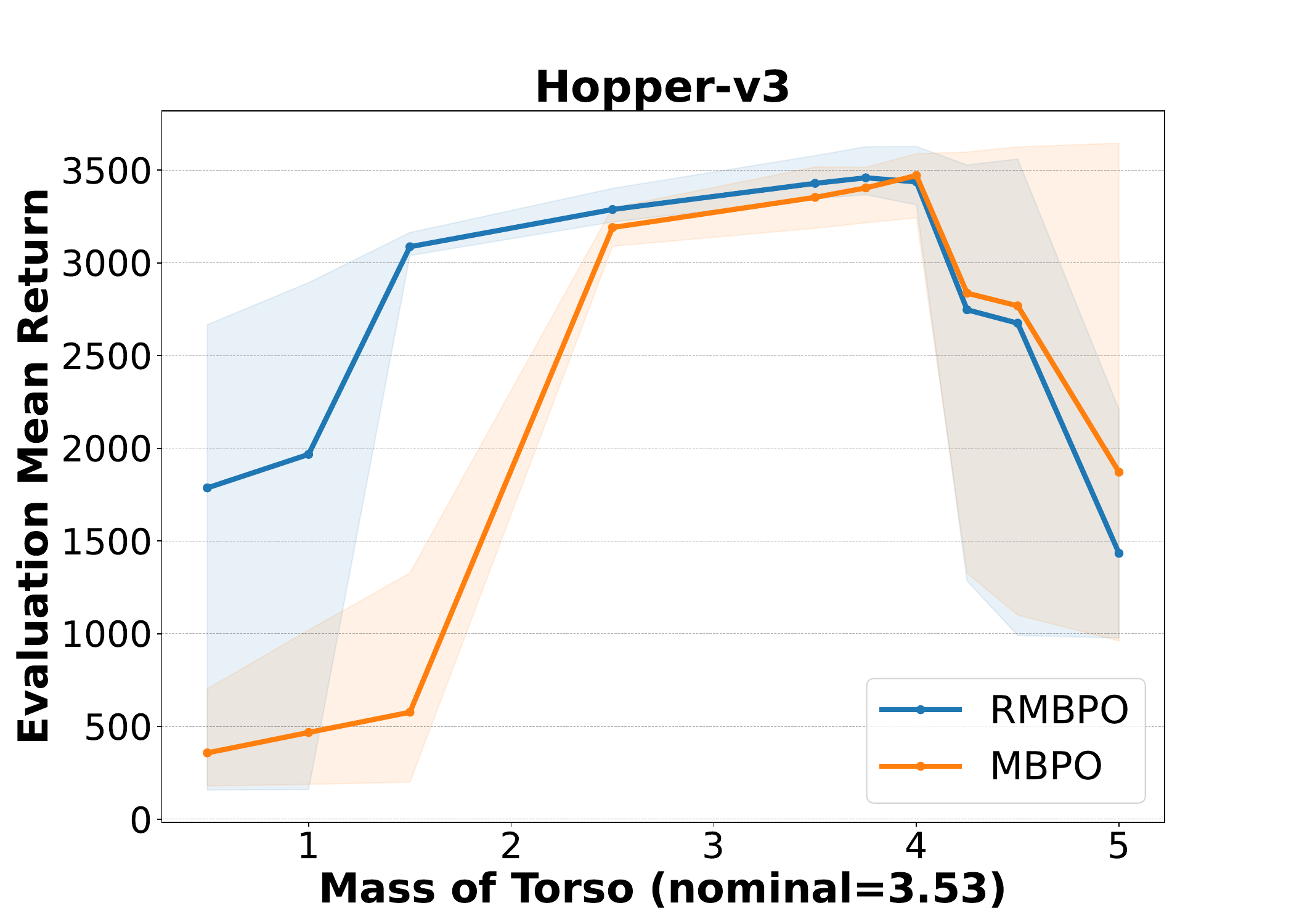}
        \caption{}
        \label{fig:hopper_robustness1}
    \end{subfigure}
    \hfill
    \begin{subfigure}{0.49\textwidth}
        \includegraphics[width=\textwidth]{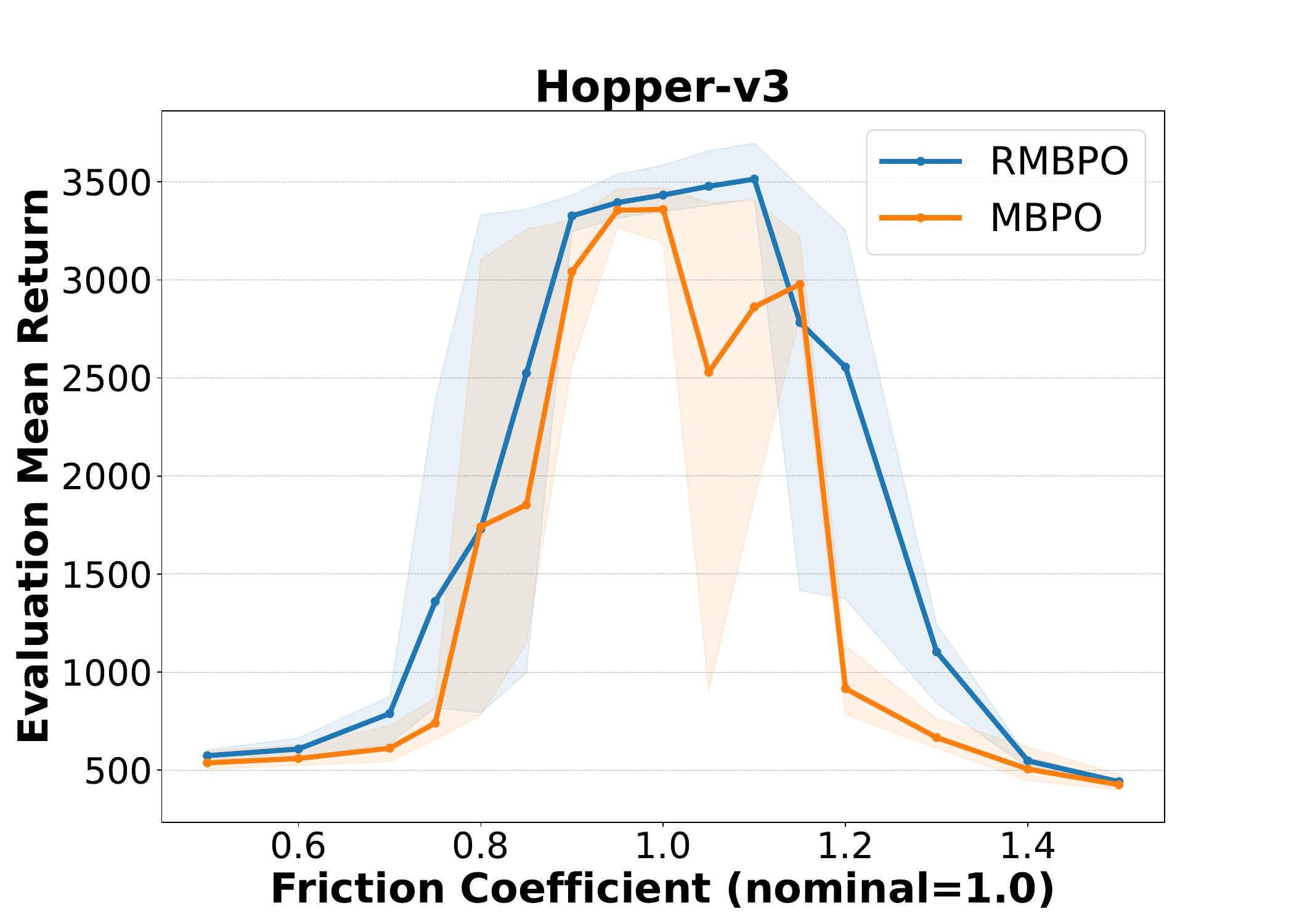}
        \caption{}
        \label{fig:hopper_robustness2}
    \end{subfigure}
    \hfill
    \begin{subfigure}{0.49\textwidth}
        \includegraphics[width=\textwidth]{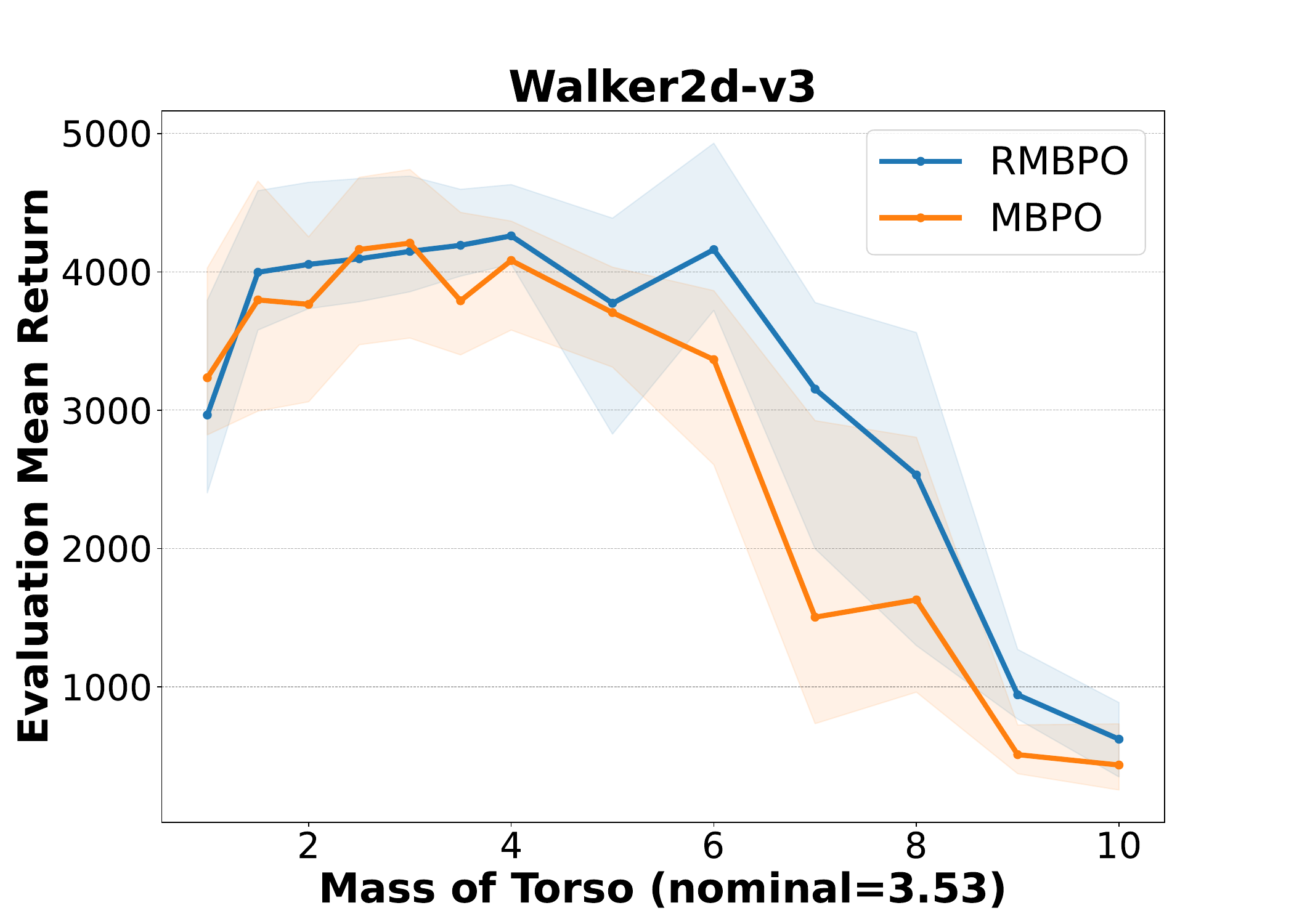}
        \caption{}
        \label{fig:walker_robustness1}
    \end{subfigure}
    \hfill
    \begin{subfigure}{0.49\textwidth}
        \includegraphics[width=\textwidth]{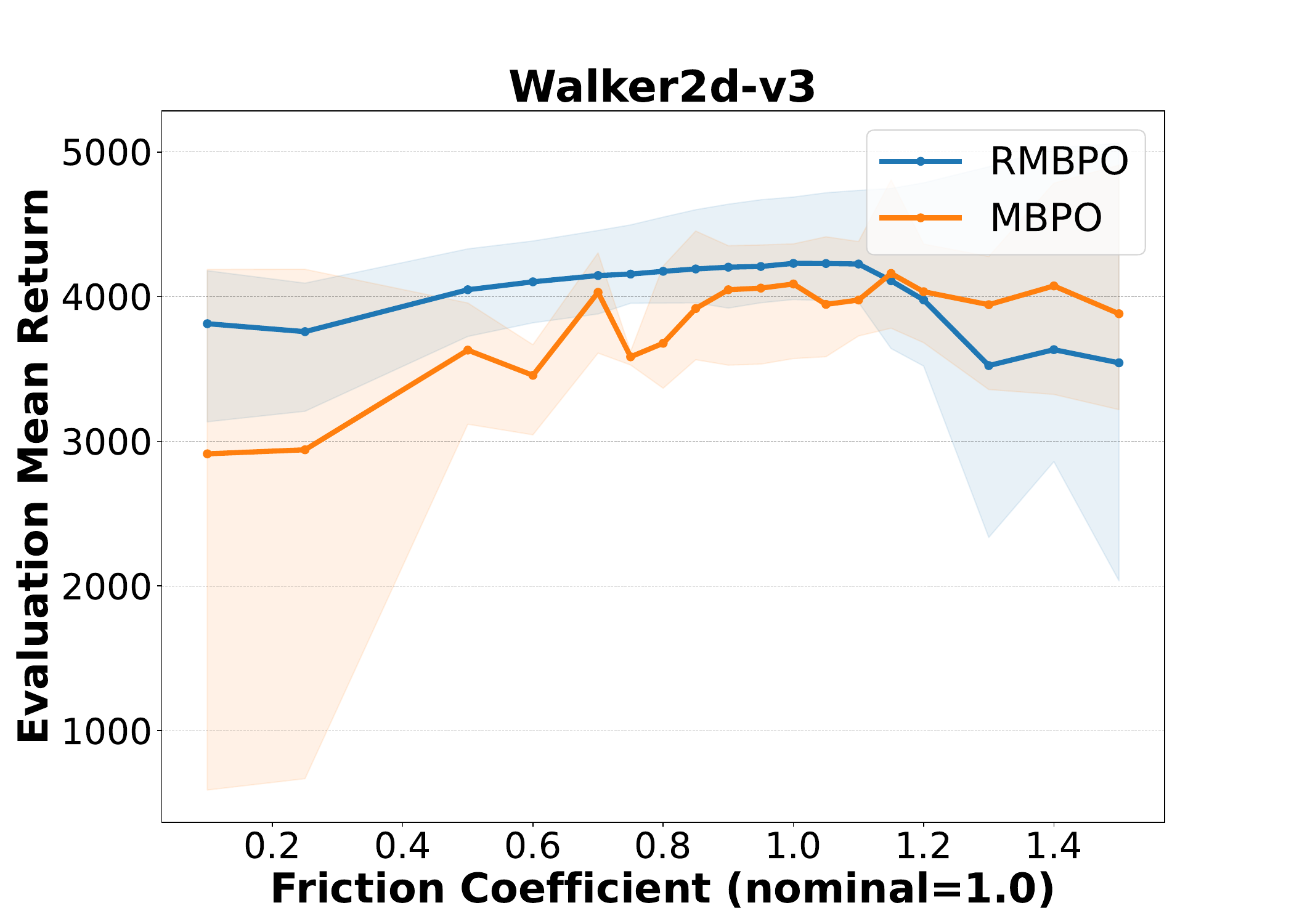}
        \caption{}
        \label{fig:walker_robustness1}
    \end{subfigure}
    \hfill
    \caption{Comparison of robustness between MBPO and RMBPO (ours). Mean return (y-axis) is plotted over the distortion of a simulation parameter during evaluation (x-axis). Results display mean over 3 seeds with the shaded region denoting the bootstrapped 95\% confidence interval between seeds.}
    \label{fig:robustness}
\end{figure}

\begin{figure}
    \centering
    \begin{subfigure}{0.49\textwidth}
        \includegraphics[width=\textwidth]{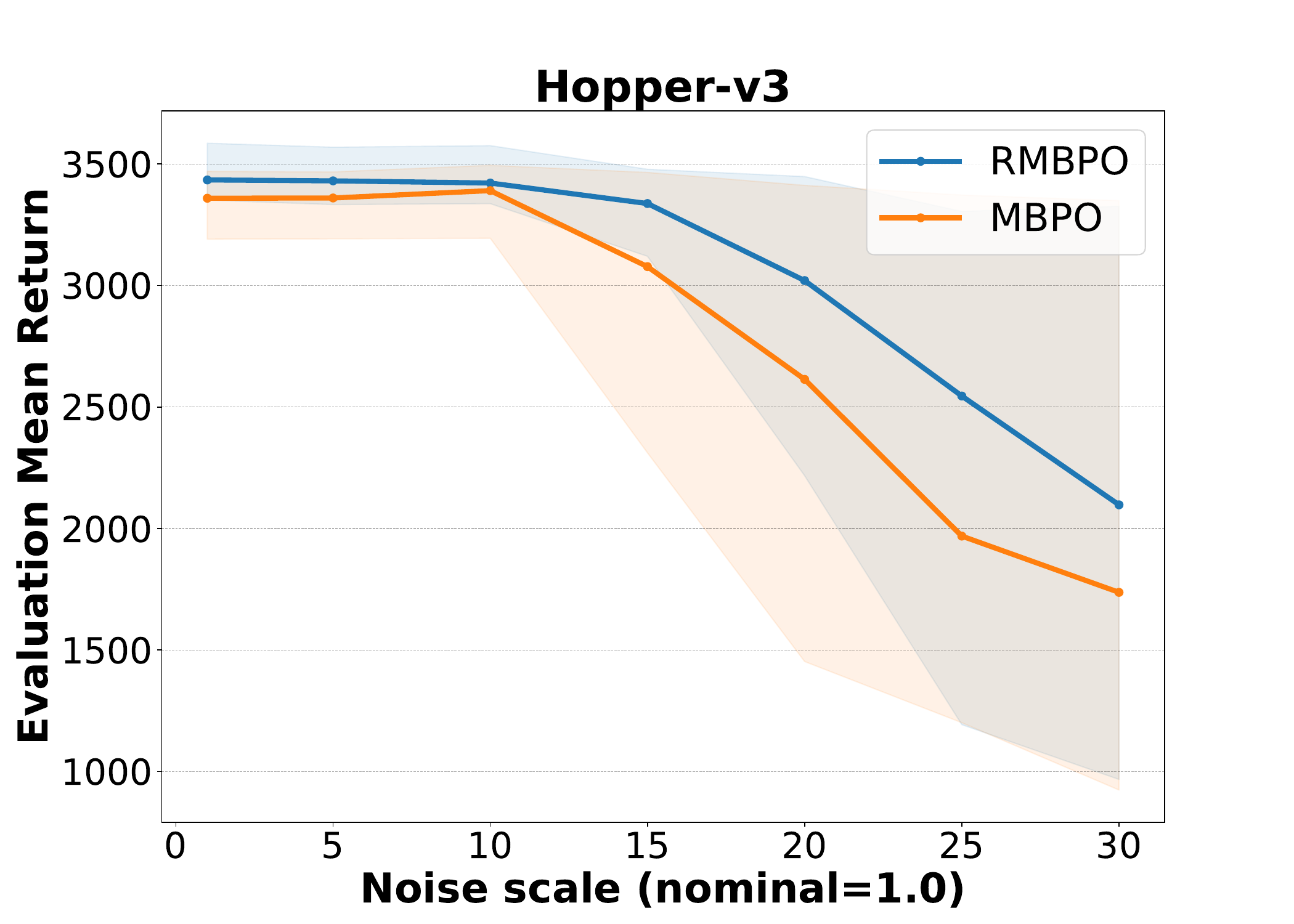}
        \caption{}
        \label{fig:hopper_noise}
    \end{subfigure}
    \hfill
    \begin{subfigure}{0.49\textwidth}
        \includegraphics[width=\textwidth]{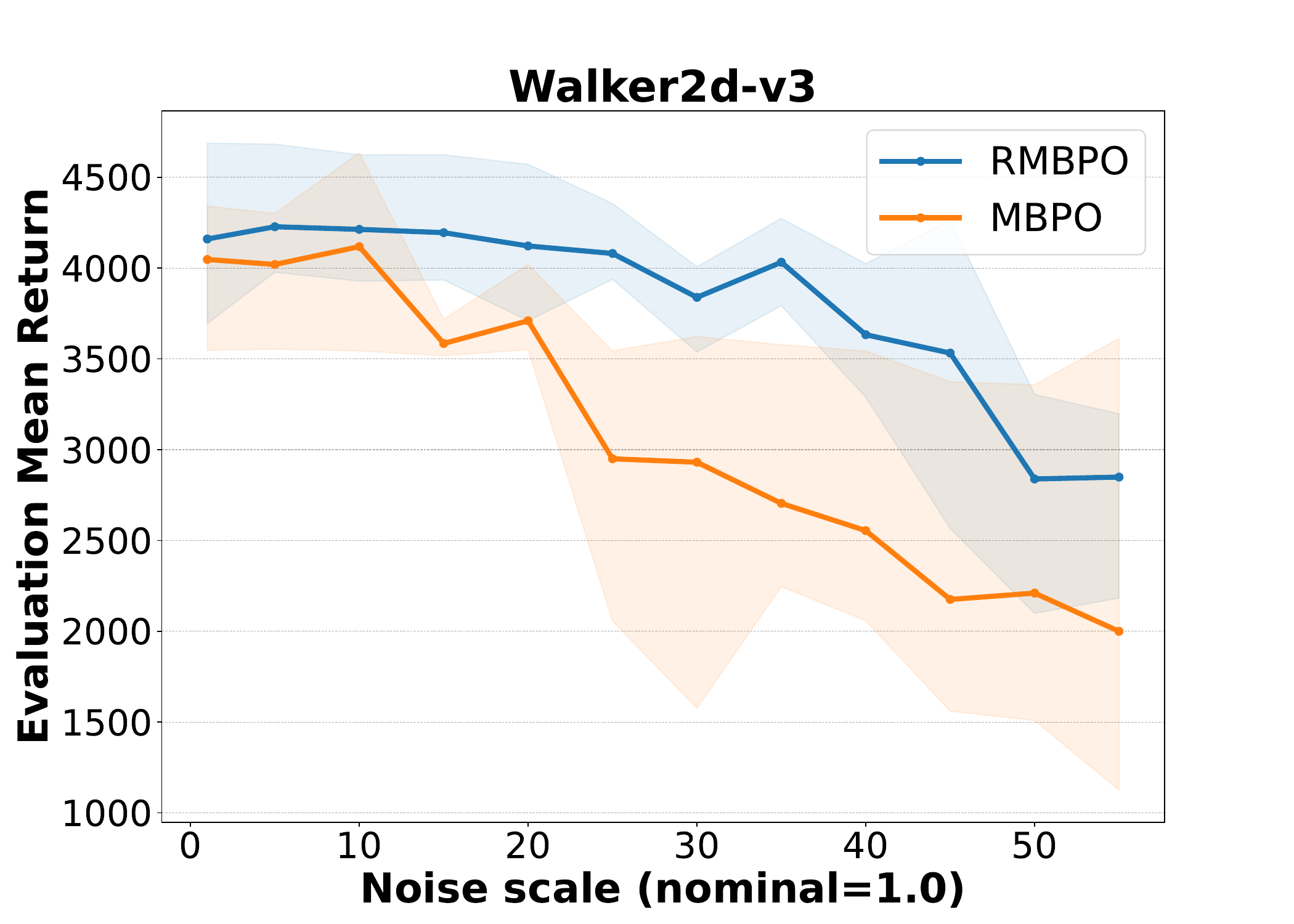}
        \caption{}
        \label{fig:walker_noise}
    \end{subfigure}
    \caption{Comparison of robustness between MBPO and RMBPO (ours). Mean return (y-axis) is plotted over the distortion of the action noise standard deviation (x-axis). Results display mean over 3 seeds with the shaded region denoting the bootstrapped 95\% confidence interval between seeds.}
    \label{fig:noise}
\end{figure}

The main goal of this section is to evaluate the hypothesis that our proposed auxiliary model aids MBRL algorithms in being more robust. To achieve our results, we train both MBPO and RMBPO in the nominal environment using 3 seeds. In accordance with \cite{agarwal2021deep}, we employ bootstrapped 95\% confidence intervals as our metric of confidence. However, in contrast to reporting the interquartile means (IQM), we report the average performance. The outlier rejection associated with IQM can yield overly optimistic results, which makes it a flawed metric when evaluating robustness. 
The results are presented in Fig. \ref{fig:robustness}, which compares the trained agents in various environments under distortion. I.e. one of the simulation parameters is modified to different values than in the nominal (training) environment.  Following \cite{pinto2017robust}, the pendulum mass is distorted in InvertedPendulum-v2, while the torso mass and friction coefficient are distorted in Hopper-v3 and Walker2d-v3. The results show that RMBPO performs better than MBPO in most of our experiments. This is the case for both the mean performance and the lower limit of the confidence interval. These results support our claim that the auxiliary model makes the policy more robust. Our results often lie between that of non-robust RL algorithms and RARL \citep{pinto2017robust}. This is an expected result since RMBPO cannot distort the simulator during training, in contrast to RARL. Additionally, in Fig. \ref{fig:noise}, we examine the robustness of RMBPO, compared to MBPO when the standard deviation of the action noise is modified. A noise scale of 1 indicates an MDP that is identical to the training environment (the nominal model), other noise scales denote multiplication factors of the standard deviation. In our results, RMBPO succeeds in being more robust to modifications of the action noise scale during evaluation.

\subsection{What is the model learning?}
\label{subsec:model_results}

\begin{figure}
    \centering
    \includegraphics[width=\textwidth]{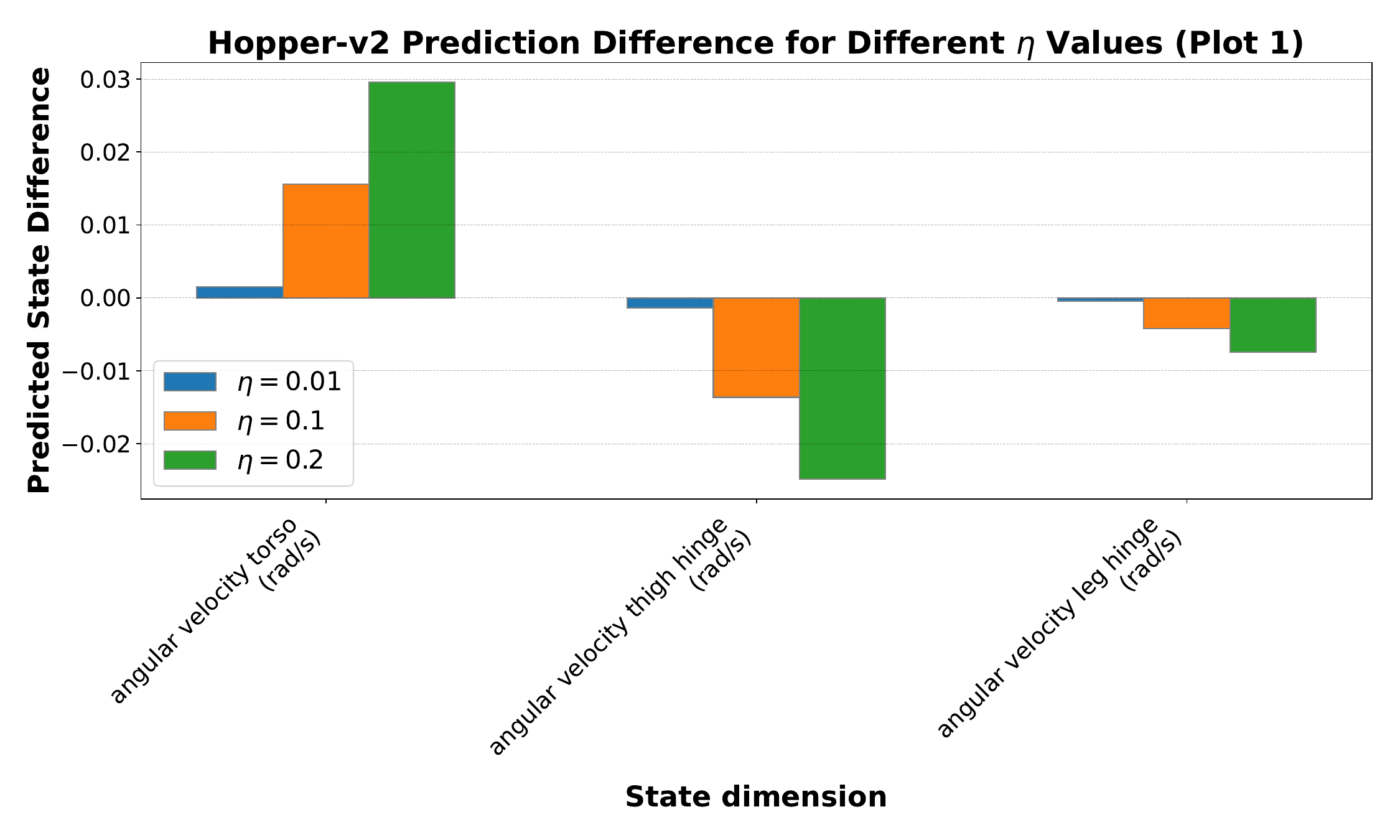}
    \caption{The difference between the nominal model and the auxiliary model. The additional state dimensions can be found in Appendix \ref{appendix:hopper_state}.}
    \label{fig:hopper_diff_1}
\end{figure}

A natural question that arises is how $g_\psi$ modifies the state transitions compared to the approximated nominal model $p_\theta \approx \bar{P}$. Also, the reader might wonder about the relationship between $\eta$ and the magnitude of these modifications. Therefore, we perform a case study on the MuJoCo Hopper-v3 environment. The observation space of this MDP consists of 11 values describing the angles and angular velocities of the joints in the robot and the position and (angular) velocity of the top of the robot. For an exhaustive list, the reader is deferred to \cite{todorov2012mujoco}. The goal of the environment is to use three rotors (in the foot, leg, and thigh) to make the robot move forward without falling. Therefore, we would expect the auxiliary model to modify the transitions in such a way that the robot moves forward more slowly and becomes more prone to falling. To examine the learned model, we display the three largest modifications (on state dimensions) that are made by the auxiliary model in Fig. \ref{fig:hopper_diff_1}. The modifications for all other state dimensions can be found in Appendix \ref{appendix:hopper_state}. It can be seen in Fig. \ref{fig:hopper_diff_1} that increasing $\eta$ consistently increases the distance of the robust predictions from the predictions of the nominal model. The three state variables that are the most influenced by the adversarial updates are the angular velocity of the torso, the thigh hinge, and the leg hinge. More importantly, it is shown that the robust model increases the angular velocity of the torso, whilst it decreases the other two. This is expected since higher mobility of the torso makes the Hopper harder to control and therefore increases the probability of it falling. The results also show that the angular velocity of the actuated parts (such as the leg and thigh) is lowered. Since these limbs are used to control the robot, this makes the system harder to control. Furthermore, Fig. \ref{fig:hopper_diff_apx_a} shows that the adversarial model lowers the x-velocity of Hopper-v3, which is associated with a lower value since the robot gets rewarded for moving forward rapidly. All these observations support the idea that the auxiliary model makes the state transitions more pessimistic, however, more examination should happen to confidently confirm this claim.

\section{Related Works}
\label{sec:related_works}
Many existing works focus on robust reinforcement learning in a tabular setting. These works include deriving a robust policy gradient \citep{wang2022policy, kumar2024policy} and providing a tractable approach to tackle non-rectangular RMDPs \citep{goyal2023robust}. In a step towards generality, \cite{wang2021online} and \cite{morimoto2005robust} consider robust reinforcement learning with function approximation on the inverted pendulum problem. As many works exist in this field, the reader is referred to \cite{moos2022robust} for more information on this topic. In the context of high-dimensional state and action spaces, \cite{pinto2017robust} propose the concept of adversarial RL for robustness. They show that an adversarial approach can improve the robustness of TRPO \citep{pmlr-v37-schulman15}. Specifically, robustness against differences between the training and testing environment is evaluated. In contrast with our work, the adversary in their methodology requires access to the simulator during training. \cite{wang2024bring} propose a methodology where multiple next states are sampled at each time step from a stochastic transition model. Subsequently, a single next state is resampled with an importance weight, based on the approximate value of that state. Similar to this work, they also consider the KL uncertainty set, however, their methodology requires a simulator where an arbitrary number of next states can be sampled at any time step. \cite{rigter2022rambo} propose an approach similar to ours, but to be robust to out-of-distribution data in offline RL. More recently, \cite{zhou2024natural} provide a model-free alternative to our work. Improved robustness against transition dynamics is demonstrated in the MuJoCo environment \citep{todorov2012mujoco} in addition to exhaustive theoretical motivation. Furthermore, \cite{queeney2024risk} consider model-free robust RL to improve safety.

\section{Conclusions and Future work}
\label{sec:conclusions}

We have proposed a novel approach for robust adversarial reinforcement learning in an online, high-dimensional setting. We have motivated the use of an auxiliary model to tackle the inner-loop optimization problem of the robust MDP formulation. This auxiliary model was then used in a practical algorithm as a modification to the model-based RL algorithm MBPO. Our experiments demonstrate the potential of the auxiliary model to improve the robustness of MBPO. Although the authors believe that this approach will work for a larger set of algorithms and environments, we leave further experimentation for future work. Other interesting areas for future work could include the addition of a secondary policy that is updated in a non-robust manner. That could ensure that the robust policy does not hinder exploration toward the optimal policy. The authors plan the extension of this work towards latent space models such as PlaNet \citep{hafner2018planet} and its more recent extensions such as DreamerV3 \citep{hafner2023mastering}.

\subsubsection*{Acknowledgments}
\label{sec:ack}
This work was supported by the Research Foundation Flanders (FWO) under Grant Number 1SHAI24N.

\bibliography{main}

\begin{thebibliography}{26}
\providecommand{\natexlab}[1]{#1}
\providecommand{\url}[1]{\texttt{#1}}
\expandafter\ifx\csname urlstyle\endcsname\relax
  \providecommand{\doi}[1]{doi: #1}\else
  \providecommand{\doi}{doi: \begingroup \urlstyle{rm}\Url}\fi

\bibitem[Agarwal et~al.(2021)Agarwal, Schwarzer, Castro, Courville, and Bellemare]{agarwal2021deep}
Rishabh Agarwal, Max Schwarzer, Pablo~Samuel Castro, Aaron~C Courville, and Marc Bellemare.
\newblock Deep reinforcement learning at the edge of the statistical precipice.
\newblock \emph{Advances in neural information processing systems}, 34:\penalty0 29304--29320, 2021.

\bibitem[Christiano et~al.(2016)Christiano, Shah, Mordatch, Schneider, Blackwell, Tobin, Abbeel, and Zaremba]{christiano2016transfer}
Paul Christiano, Zain Shah, Igor Mordatch, Jonas Schneider, Trevor Blackwell, Joshua Tobin, Pieter Abbeel, and Wojciech Zaremba.
\newblock Transfer from simulation to real world through learning deep inverse dynamics model.
\newblock \emph{arXiv preprint arXiv:1610.03518}, 2016.

\bibitem[Feng(2021)]{unstableBaselines}
Xu~Feng.
\newblock Unstable baselines.
\newblock \url{https://github.com/x35f/unstable_baselines}, 2021.

\bibitem[Goyal \& Grand-Clement(2023)Goyal and Grand-Clement]{goyal2023robust}
Vineet Goyal and Julien Grand-Clement.
\newblock Robust markov decision processes: Beyond rectangularity.
\newblock \emph{Mathematics of Operations Research}, 48\penalty0 (1):\penalty0 203--226, 2023.

\bibitem[Haarnoja et~al.(2018)Haarnoja, Zhou, Abbeel, and Levine]{haarnoja2018soft}
Tuomas Haarnoja, Aurick Zhou, Pieter Abbeel, and Sergey Levine.
\newblock Soft actor-critic: Off-policy maximum entropy deep reinforcement learning with a stochastic actor.
\newblock In \emph{International conference on machine learning}, pp.\  1861--1870. PMLR, 2018.

\bibitem[Hafner et~al.(2018)Hafner, Lillicrap, Fischer, Villegas, Ha, Lee, and Davidson]{hafner2018planet}
Danijar Hafner, Timothy Lillicrap, Ian Fischer, Ruben Villegas, David Ha, Honglak Lee, and James Davidson.
\newblock Learning latent dynamics for planning from pixels.
\newblock \emph{arXiv preprint arXiv:1811.04551}, 2018.

\bibitem[Hafner et~al.(2023)Hafner, Pasukonis, Ba, and Lillicrap]{hafner2023mastering}
Danijar Hafner, Jurgis Pasukonis, Jimmy Ba, and Timothy Lillicrap.
\newblock Mastering diverse domains through world models.
\newblock \emph{arXiv preprint arXiv:2301.04104}, 2023.

\bibitem[Hu \& Hong(2013)Hu and Hong]{hu2013kullback}
Zhaolin Hu and L~Jeff Hong.
\newblock Kullback-leibler divergence constrained distributionally robust optimization.
\newblock \emph{Available at Optimization Online}, 1\penalty0 (2):\penalty0 9, 2013.

\bibitem[Janner et~al.(2019)Janner, Fu, Zhang, and Levine]{janner2019trust}
Michael Janner, Justin Fu, Marvin Zhang, and Sergey Levine.
\newblock When to trust your model: Model-based policy optimization.
\newblock \emph{Advances in neural information processing systems}, 32, 2019.

\bibitem[Kumar et~al.(2024)Kumar, Derman, Geist, Levy, and Mannor]{kumar2024policy}
Navdeep Kumar, Esther Derman, Matthieu Geist, Kfir~Y Levy, and Shie Mannor.
\newblock Policy gradient for rectangular robust markov decision processes.
\newblock \emph{Advances in Neural Information Processing Systems}, 36, 2024.

\bibitem[Moerland et~al.(2023)Moerland, Broekens, Plaat, Jonker, et~al.]{moerland2023model}
Thomas~M Moerland, Joost Broekens, Aske Plaat, Catholijn~M Jonker, et~al.
\newblock Model-based reinforcement learning: A survey.
\newblock \emph{Foundations and Trends{\textregistered} in Machine Learning}, 16\penalty0 (1):\penalty0 1--118, 2023.

\bibitem[Moos et~al.(2022)Moos, Hansel, Abdulsamad, Stark, Clever, and Peters]{moos2022robust}
Janosch Moos, Kay Hansel, Hany Abdulsamad, Svenja Stark, Debora Clever, and Jan Peters.
\newblock Robust reinforcement learning: A review of foundations and recent advances.
\newblock \emph{Machine Learning and Knowledge Extraction}, 4\penalty0 (1):\penalty0 276--315, 2022.

\bibitem[Morimoto \& Doya(2005)Morimoto and Doya]{morimoto2005robust}
Jun Morimoto and Kenji Doya.
\newblock Robust reinforcement learning.
\newblock \emph{Neural computation}, 17\penalty0 (2):\penalty0 335--359, 2005.

\bibitem[Pinto et~al.(2017)Pinto, Davidson, Sukthankar, and Gupta]{pinto2017robust}
Lerrel Pinto, James Davidson, Rahul Sukthankar, and Abhinav Gupta.
\newblock Robust adversarial reinforcement learning.
\newblock In \emph{International Conference on Machine Learning}, pp.\  2817--2826. PMLR, 2017.

\bibitem[Queeney \& Benosman(2024)Queeney and Benosman]{queeney2024risk}
James Queeney and Mouhacine Benosman.
\newblock Risk-averse model uncertainty for distributionally robust safe reinforcement learning.
\newblock \emph{Advances in Neural Information Processing Systems}, 36, 2024.

\bibitem[Rigter et~al.(2022)Rigter, Lacerda, and Hawes]{rigter2022rambo}
Marc Rigter, Bruno Lacerda, and Nick Hawes.
\newblock Rambo-rl: Robust adversarial model-based offline reinforcement learning.
\newblock \emph{Advances in neural information processing systems}, 35:\penalty0 16082--16097, 2022.

\bibitem[Rusu et~al.(2017)Rusu, Ve{\v{c}}er{\'\i}k, Roth{\"o}rl, Heess, Pascanu, and Hadsell]{rusu2017sim}
Andrei~A Rusu, Matej Ve{\v{c}}er{\'\i}k, Thomas Roth{\"o}rl, Nicolas Heess, Razvan Pascanu, and Raia Hadsell.
\newblock Sim-to-real robot learning from pixels with progressive nets.
\newblock In \emph{Conference on robot learning}, pp.\  262--270. PMLR, 2017.

\bibitem[Schulman et~al.(2015)Schulman, Levine, Abbeel, Jordan, and Moritz]{pmlr-v37-schulman15}
John Schulman, Sergey Levine, Pieter Abbeel, Michael Jordan, and Philipp Moritz.
\newblock Trust region policy optimization.
\newblock In Francis Bach and David Blei (eds.), \emph{Proceedings of the 32nd International Conference on Machine Learning}, volume~37 of \emph{Proceedings of Machine Learning Research}, pp.\  1889--1897, Lille, France, 07--09 Jul 2015. PMLR.
\newblock URL \url{https://proceedings.mlr.press/v37/schulman15.html}.

\bibitem[Todorov et~al.(2012)Todorov, Erez, and Tassa]{todorov2012mujoco}
Emanuel Todorov, Tom Erez, and Yuval Tassa.
\newblock Mujoco: A physics engine for model-based control.
\newblock In \emph{2012 IEEE/RSJ International Conference on Intelligent Robots and Systems}, pp.\  5026--5033, 2012.
\newblock \doi{10.1109/IROS.2012.6386109}.

\bibitem[Virtanen et~al.(2020)Virtanen, Gommers, Oliphant, Haberland, Reddy, Cournapeau, Burovski, Peterson, Weckesser, Bright, {van der Walt}, Brett, Wilson, Millman, Mayorov, Nelson, Jones, Kern, Larson, Carey, Polat, Feng, Moore, {VanderPlas}, Laxalde, Perktold, Cimrman, Henriksen, Quintero, Harris, Archibald, Ribeiro, Pedregosa, {van Mulbregt}, and {SciPy 1.0 Contributors}]{2020SciPy-NMeth}
Pauli Virtanen, Ralf Gommers, Travis~E. Oliphant, Matt Haberland, Tyler Reddy, David Cournapeau, Evgeni Burovski, Pearu Peterson, Warren Weckesser, Jonathan Bright, St{\'e}fan~J. {van der Walt}, Matthew Brett, Joshua Wilson, K.~Jarrod Millman, Nikolay Mayorov, Andrew R.~J. Nelson, Eric Jones, Robert Kern, Eric Larson, C~J Carey, {\.I}lhan Polat, Yu~Feng, Eric~W. Moore, Jake {VanderPlas}, Denis Laxalde, Josef Perktold, Robert Cimrman, Ian Henriksen, E.~A. Quintero, Charles~R. Harris, Anne~M. Archibald, Ant{\^o}nio~H. Ribeiro, Fabian Pedregosa, Paul {van Mulbregt}, and {SciPy 1.0 Contributors}.
\newblock {{SciPy} 1.0: Fundamental Algorithms for Scientific Computing in Python}.
\newblock \emph{Nature Methods}, 17:\penalty0 261--272, 2020.
\newblock \doi{10.1038/s41592-019-0686-2}.

\bibitem[Wang et~al.(2024)Wang, Gadot, Kumar, Levy, and Mannor]{wang2024bring}
Kaixin Wang, Uri Gadot, Navdeep Kumar, Kfir Levy, and Shie Mannor.
\newblock Bring your own (non-robust) algorithm to solve robust mdps by estimating the worst kernel, 2024.

\bibitem[Wang \& Zou(2021)Wang and Zou]{wang2021online}
Yue Wang and Shaofeng Zou.
\newblock Online robust reinforcement learning with model uncertainty.
\newblock \emph{Advances in Neural Information Processing Systems}, 34:\penalty0 7193--7206, 2021.

\bibitem[Wang \& Zou(2022)Wang and Zou]{wang2022policy}
Yue Wang and Shaofeng Zou.
\newblock Policy gradient method for robust reinforcement learning.
\newblock In \emph{International conference on machine learning}, pp.\  23484--23526. PMLR, 2022.

\bibitem[Wiesemann et~al.(2013)Wiesemann, Kuhn, and Rustem]{wiesemann2013robust}
Wolfram Wiesemann, Daniel Kuhn, and Ber{\c{c}} Rustem.
\newblock Robust markov decision processes.
\newblock \emph{Mathematics of Operations Research}, 38\penalty0 (1):\penalty0 153--183, 2013.

\bibitem[Zhao et~al.(2020)Zhao, Queralta, and Westerlund]{zhao2020sim}
Wenshuai Zhao, Jorge~Pe{\~n}a Queralta, and Tomi Westerlund.
\newblock Sim-to-real transfer in deep reinforcement learning for robotics: a survey.
\newblock In \emph{2020 IEEE symposium series on computational intelligence (SSCI)}, pp.\  737--744. IEEE, 2020.

\bibitem[Zhou et~al.(2024)Zhou, Liu, Cheng, Kalathil, Kumar, and Tian]{zhou2024natural}
Ruida Zhou, Tao Liu, Min Cheng, Dileep Kalathil, PR~Kumar, and Chao Tian.
\newblock Natural actor-critic for robust reinforcement learning with function approximation.
\newblock \emph{Advances in neural information processing systems}, 36, 2024.

\end{thebibliography}
\bibliographystyle{rlc}

\appendix

\section{Hyperparamters}
\label{appendix:hyperparams}

\begin{table}[htbp]
    \begin{center}
        \begin{tabular}{llll}
            \multicolumn{1}{l}{\bf Hyperparameter}  &\multicolumn{1}{l}{\bf InvertedPendulum-v2} &\multicolumn{1}{l}{\bf Hopper-v3} &\multicolumn{1}{l}{\bf Walker2d-v3}
            \\ \hline \\
            $\eta$         &0.3 &0.004 &0.001 \\
            $\lambda_a$         &1e-4 &1e-4 &1e-4 \\
            Total environment steps &15e3 &125e3 &230e3 \\
        \end{tabular}
    \end{center}
    \caption{Hyperparameters}
    \label{tab:exampleTable}
\end{table}

We hypothesize that the optimal value of $\eta$ is related to the cardinality of the state space of the MDP, however, we leave further investigation for future work. The pessimistic model learning rate ($\lambda_a$) is set to 1/10 of the normal MBPO model learning rate, this significantly reduces variance on the return during training. Note that we use the same amount of environment steps as MBPO in InvertedPendulum and Hopper, but we use a lower amount in Walker2d (230k compared to 300k), this helped to significantly reduce experiment duration and computational cost. In future work, we aim to evaluate our methodology on more environments for a longer amount of steps.

All other hyperparameters remain identical to MBPO \citep{janner2019trust}, the auxiliary model $g_{\psi}$ also has the same architecture as the MBPO world model.  

\section{Implementation details}
\label{appendix:details}
Following related work \citep{zhou2024natural}, we slightly modify the mentioned MuJoCo environments by adding Gaussian noise to the action: $a_t \leftarrow a_t + \mathcal{N}(0, 5e-3)$. Only in Pendulum-v2, we use a significantly higher noise variance of 0.2. Since this action noise is invisible to the agent, it introduces stochasticity in the MDP. Inspired by the existing MBPO world model, we standardize the outputs of $p_{\theta}$ before providing them as inputs to $g_{\psi}$, this showed incremental stability improvements in some environments. As proposed in appendix A.1 by \cite{rigter2022rambo}, we subtract $V^{\theta, \psi}_{\phi}(s)$ as a baseline from the return in Eq.~\ref{eqn:aux_loss}, this does not influence the expectation of the gradient but significantly reduces its variance. Note that MBPO does not employ a value network directly, but this does not pose an issue since on-policy samples from the Q-network will provide the same expected gradient, given a large enough minibatch size. 

Our implementation is based upon the Unstable Baselines Python library \citep{unstableBaselines}. We preferred this implementation because of its clarity, however, we experimentally verified that Unstable Baselines reached the same performance as the original open-source MBPO code. All hyperparameters of MBPO remain identical to \cite{janner2019trust}. For calculating the bootstrapped confidence intervals, we used the implementation provided by SciPy \citep{2020SciPy-NMeth}. Experiments were run on a Ubuntu20.04 (Docker) machine with a single NVIDIA v100 GPU, two CPU cores, and 10GB of memory.

\section{Reproducibility}
To improve reproducibility, we provide all implementation details that are not mentioned in the main body of the paper in Appendix \ref{appendix:details}. Furthermore, we provide the trained weights of the learned policies as supplementary materials, together with the modified environments and an evaluation script \footnote{\url{https://github.com/rmbpo-eval/rmbpo-eval}}. This allows for a clear comparison with our research. We chose to evaluate by distorting the same model parameters as \cite{pinto2017robust} for two reasons: a) to add perspective to the results and ease future benchmarking in the community, and b) to avoid cherry-picking the best conditions for RMBPO. To improve the trustworthiness of our results, we share the same 3 seeds throughout all of our experiments, to avoid picking the best seeds per environment. The authors are not able to release source code at the time of submission of this paper, however, the reader is encouraged to contact the first author of this work with any related questions.

\newpage

\section{Additional state difference graphs Hopper-v3}
\label{appendix:hopper_state}

\begin{figure}[h]
    \centering
    \subfloat[Medium state differences]{%
      \includegraphics[clip,width=\textwidth]{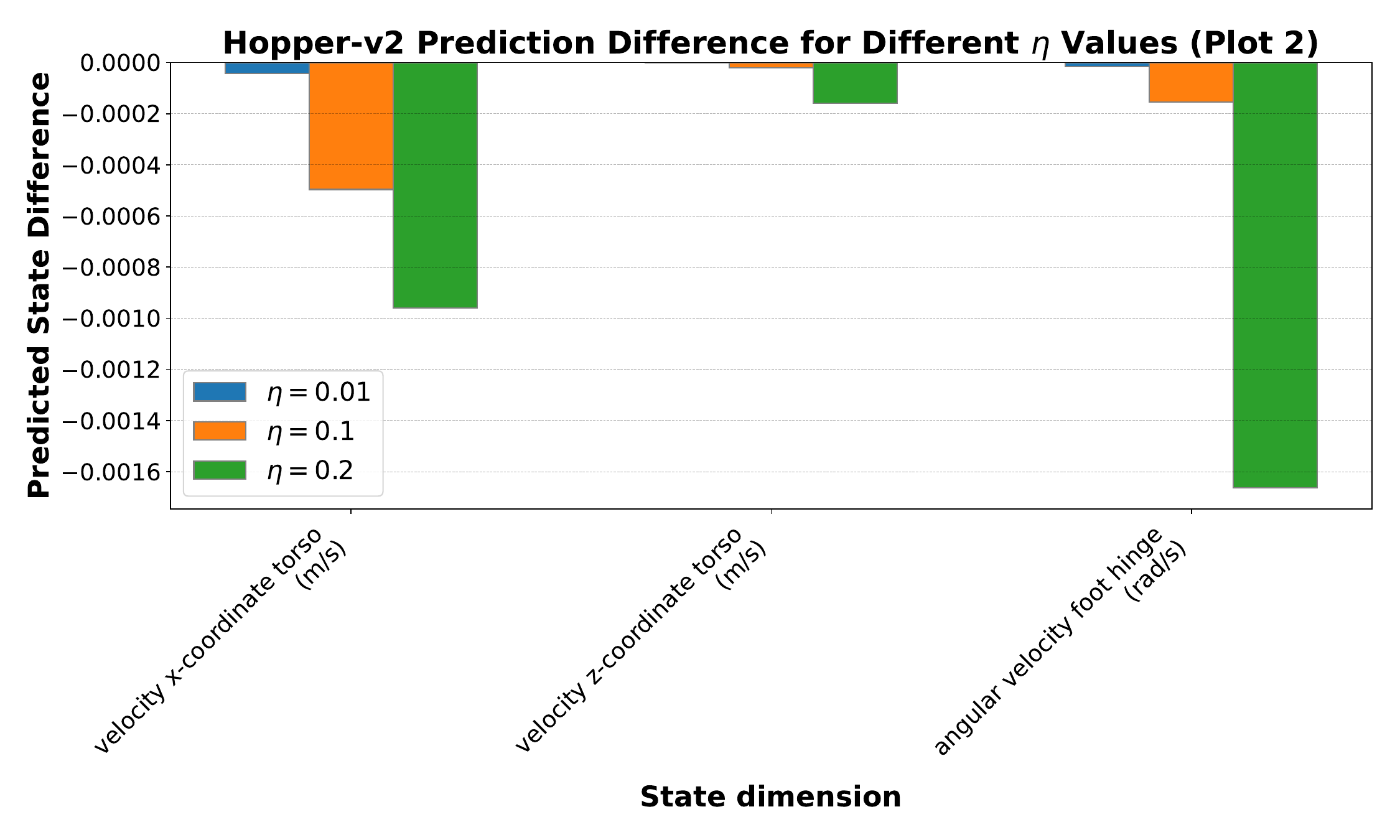}%
      \label{fig:hopper_diff_apx_a}
    }
    
    \subfloat[Small state differences]{%
      \includegraphics[clip,width=\textwidth]{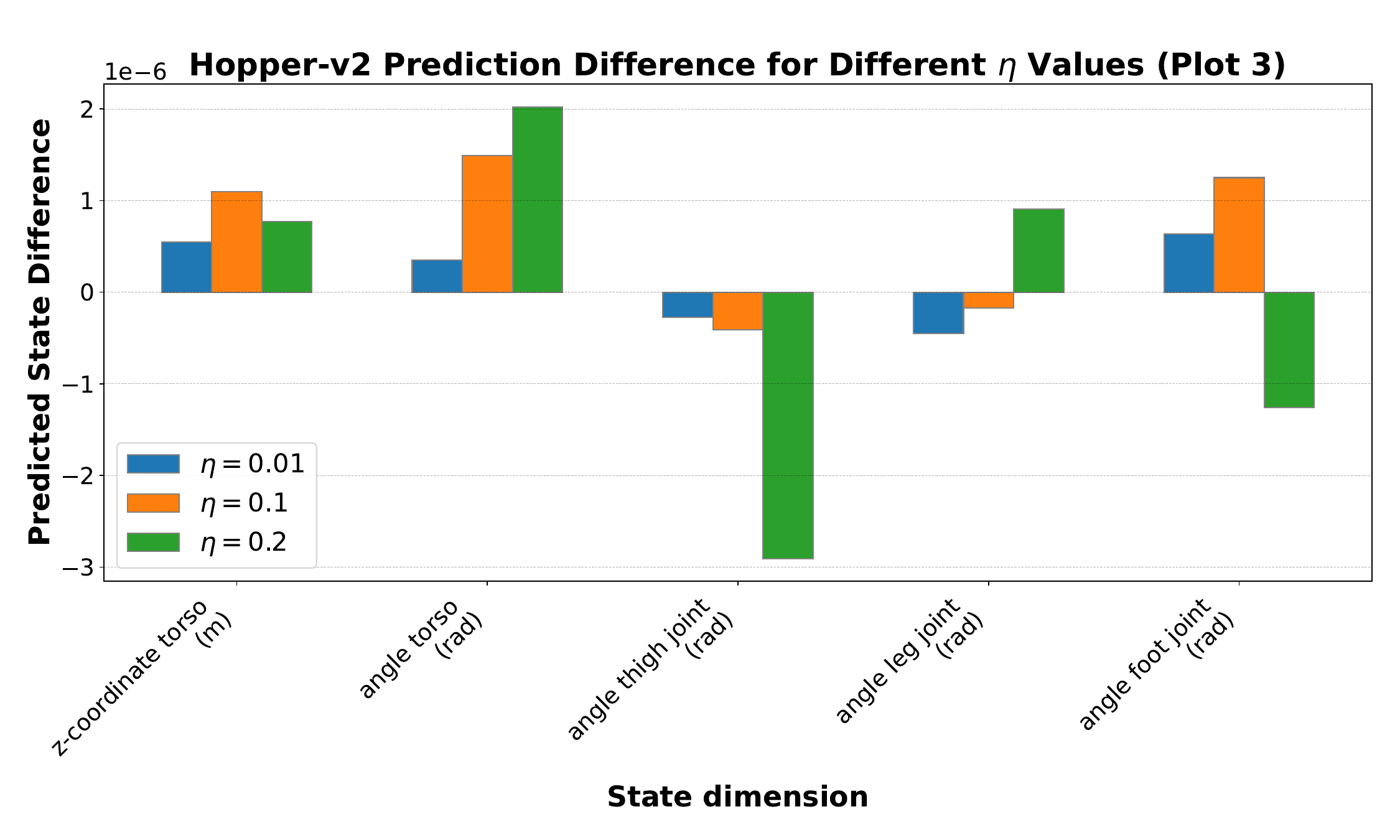}%
       \label{fig:hopper_diff_apx_b}
    }
    \caption{Additional Hopper-v3 state differences.}
    \label{fig:hopper_diff_apx}
\end{figure}

\end{document}